\documentclass[10pt,twocolumn,letterpaper]{article}

\usepackage{wacv}
\usepackage{times}
\usepackage{epsfig}
\usepackage{graphicx}
\usepackage{amsmath}

\usepackage{comment}
\usepackage{epstopdf}
\usepackage{graphics}
\usepackage{subfigure}
\usepackage{xspace}
\usepackage{multirow}
\usepackage{float}
\usepackage{tabularx}
\usepackage{gensymb}
\usepackage{anyfontsize}
\usepackage{caption}
\usepackage{enumitem}  
\usepackage{xcolor}
\usepackage{amssymb} 
\usepackage{pifont}
\usepackage{booktabs}
\usepackage{colortbl}
\usepackage{balance}
\usepackage[accsupp]{axessibility} 

\def\wacvPaperID{790} 

\wacvfinalcopy 

\ifwacvfinal
\fi

\ifwacvfinal
\usepackage[breaklinks=true,bookmarks=false]{hyperref}
\else
\usepackage[pagebackref=true,breaklinks=true,colorlinks,bookmarks=false]{hyperref}
\fi

\begin{document}

\title{Multi-View Fusion of Sensor Data for Improved \\Perception and Prediction in Autonomous Driving}


\renewcommand\footnotemark{}
\renewcommand\footnoterule{}

\author{ 
  Sudeep Fadadu$^{*}$, Shreyash Pandey$^{*}$, Darshan Hegde, Yi Shi,\\ 
  Fang-Chieh Chou, Nemanja Djuric, Carlos Vallespi-Gonzalez \\
  \thanks{
    Authors are with Aurora Innovation, work done while at Uber ATG.
    E-mail contacts: {\tt\small \{sfadadu, spandey, darshan.hegde, yishi, fchou,  ndjuric, cvallespi\}@aurora.tech}
  }
  \thanks{$^{*}$ Authors contributed equally.}
}

\maketitle

\ifwacvfinal
\thispagestyle{empty}
\fi

\begin{abstract}
We present an end-to-end method for object detection and trajectory prediction utilizing multi-view representations of LiDAR returns and camera images. 
In this work, we recognize the strengths and weaknesses of different view representations, and we propose an efficient and generic fusing method that aggregates benefits from all views.
Our model builds on a state-of-the-art Bird's-Eye View (BEV) network that fuses voxelized features from a sequence of historical LiDAR data as well as rasterized high-definition map to perform detection and prediction tasks.
We extend this model with additional LiDAR Range-View (RV) features that use the raw LiDAR information in its native, non-quantized representation. 
The RV feature map is projected into BEV and fused with the BEV features computed from LiDAR and high-definition map. 
The fused features are then further processed to output the final detections and trajectories, within a single end-to-end trainable network.
In addition, the RV fusion of LiDAR and camera is performed in a straightforward and computationally efficient manner using this framework.
The proposed multi-view fusion approach improves the state-of-the-art on proprietary large-scale real-world data collected by a fleet of self-driving vehicles, as well as on the public nuScenes data set with minimal increases on the computational cost.

\end{abstract}

\section{INTRODUCTION} 
\label{sect:introduction}

Self-driving vehicle (SDV) is a highly involved system, relying on several input sensors (such as LiDARs, radars, and cameras) to detect objects in its vicinity, and predict their future motion and the motion uncertainty \cite{Bertha2015}.
Numerous works were recently proposed to address object detection from raw sensor data \cite{Meyer_2019_CVPR_Workshops,meyer2019lasernet}, as well as their motion prediction where objects are taken as inputs \cite{cui2019multimodal,djuric2020}.
Here, the two tasks are solved separately and in a sequence, with the detection and prediction modules trained in isolation.
However, such multi-stage systems may result in added system latency due to lack of feature sharing, and also suffer from cascading errors.
To address this issue, joint models performing the two tasks as part of a single end-to-end system were proposed \cite{casas2018intentnet}, where gradients can flow from output trajectories to detections and further back to the sensor data.

Most recent works on end-to-end models operate in the Bird's-Eye View (BEV) \cite{casas2018intentnet,djuric2020multixnet}, projecting the raw LiDAR data into a top-down grid centered on the SDV.
BEV encoding of LiDAR returns has several advantages that make the object detection and motion forecasting task easier to learn. 
One such advantage is that the size of objects remains constant regardless of range, which provides a strong prior to the output space and simplifies the problem. 
Other advantages include effective fusion of historical LiDAR data, as well as efficient fusion of high-definition (HD) map features \cite{djuric2020multixnet}.
However, this representation discretizes the LiDAR data into voxels, losing fine-grained information necessary to detect smaller objects such as pedestrians and bicyclists. 

On the other hand, Range-View (RV) methods operate in LiDAR's native, dense representation, providing full access to the non-quantized sensor information.
This property was shown to provide both efficiency benefits and strong detection performance, especially for smaller objects like pedestrians and bicyclists \cite{meyer2019lasernet}.
RV representation is also arguably better suited for efficiently fusing information from sensors that natively capture data in RV, like LiDAR and camera \cite{Meyer_2019_CVPR_Workshops}. 
However, these approaches operate on RV inputs while outputting detections in a BEV space, implying that the model needs to learn the transformation from RV to BEV. 
Moreover, they also have to handle variations in perceived object size with range. 
These issues make the problem more complex, and the model thus requires a larger data set in order to be competitive with the BEV-based methods.
Recently, methods such as \cite{meyer2020laserflow} also showed that fusing historical LiDAR data in RV is challenging due to distortions that arise due to shift in the center of spherical projection.

In this work, we propose an architecture that combines the benefits of the above-mentioned BEV and RV methods under a unified framework.
In particular, we introduce a novel and improved data fusion scheme that projects LiDAR data into both BEV and RV representations, where it is combined with HD map data and camera, respectively.
Then, the two views are processed separately before being fused in a common BEV frame by using LiDAR matching, followed by joint detection and motion prediction. 
Our contributions are summarized below:
\begin{itemize}
    \item We propose multi-view encoding and processing of LiDAR data separately in BEV and RV frames, before fusing the two views in a common BEV feature space.
    \item We propose computationally efficient sensor fusion of the camera RGB data with LiDAR in the RV frame, before projecting the learned features to the BEV frame.
    \item We evaluate the fusion approach by applying it to the state-of-the-art MultiXNet \cite{djuric2020multixnet}. Nevertheless the proposed multi-view fusion is quite general and can be  applied to improve other BEV- and RV-based methods.
    \item We present an ablation study and compare the method to the state-of-the-art on proprietary and open-sourced data, indicating improvements in both object detection and motion prediction tasks.
\end{itemize}
\section{RELATED WORK}
\label{sect:related_work}

Accurate 3D object detection using LiDAR point clouds is a key technology in SDV development \cite{guo2020deep}. 
To apply deep neural networks on point cloud data, it needs to be converted into appropriate feature representations that can be ingested by the deep models. 
A popular approach is to treat the points as an unordered set of point-wise feature vectors, first proposed by PointNet \cite{Charles_2017}, and applied to object detection in F-PointNet \cite{qi2018frustum} and StarNet \cite{ngiam2019starnet}. 
Another common method is to voxelize the point cloud onto BEV grid cells and apply CNN on the BEV feature map, as done in PIXOR for object detection \cite{yang2018pixor}. 
HDNET \cite{pmlr-v87-yang18b} extends PIXOR with additional HD map inputs. VoxelNet \cite{zhou2018voxelnet} and PointPillar \cite{lang2019pointpillars} use point-based feature extraction on each BEV grid cells before applying CNN. 
Lastly, LaserNet \cite{meyer2019lasernet} (and other similar approaches like \cite{LueFan2021rv}) proposes an RV representation of LiDAR points, by unwarping the cylindrical LiDAR sweep onto a 2D feature map, where each pixel represents features of a single LiDAR point.

Most of the detection methods, as discussed above, use only one feature representation of LiDAR. 
Some recent research explore the idea of combining multiple LiDAR feature representations, known as the Multi-View (MV) approach. 
MV3D \cite{chen2017multi} uses a BEV-based network to perform region proposal, applies Region of Interest (ROI) pooling on BEV, RV, and image feature map to extract 1D feature vector for each view, and fuses the 1D features of different views before the final detection heads. 
MVF \cite{zhou2019end} first voxelizes the point cloud in two different views: BEV and perspective view, extracts point-wise embedding in the two views separately to augment the point-wise feature, then applies PointPillar-style BEV network to generate detection. 
Each of the methods discussed above has certain weaknesses. 
MV3D uses a large network for each view and only performs late MV fusion after ROI cropping.
MVF uses expensive additional networks to obtain point-wise embedding in multiple views before applying the backbone BEV network. 
In comparison, our proposed method extracts informative RV features with a lightweight network and fuses to existing state-of-the-art BEV architecture before object detection is performed, giving significant performance improvements at a small computation cost.

A related field of research is LiDAR-camera fusion in object detection; \cite{feng2020deep} gives an overview of sensor fusion in autonomous driving.
Camera images are natively presented in the 2D front view, distinct from the common BEV representation for LiDAR data. 
In addition, the camera images do not give direct depth measurements on each pixel, so one cannot trivially project the camera pixels onto BEV grids. 
AVOD \cite{ku2018joint} and MV3D \cite{chen2017multi} fuse camera and LiDAR features after second-stage feature cropping. 
F-PointNet \cite{qi2018frustum} applies 2D object detection method on the camera image, then uses PointNet on LiDAR points falling into the frustum of each 2D detection to generate 3D detections. 
Continuous fusion (ContFuse) \cite{liang2018deep} finds the nearest LiDAR points for each BEV grid cell, and projects these 3D points onto camera image to extract corresponding image features.
MMF \cite{liang2019multi} combines several ideas: it fuses camera features to BEV features at multiple scales using ContFuse, augments ContFuse with additional pseudo-LiDAR points from an image depth estimator, and performs second-stage fusion by concatenating feature map crops from BEV and camera view.
Lastly, recently proposed LaserNet++ \cite{Meyer_2019_CVPR_Workshops} extended LaserNet with image fusion. 
Since the RV feature in LaserNet is in the same front view as the camera, LaserNet++ simply projects the camera feature onto the RV feature map, and the authors show improvements over state-of-the-art detection and LiDAR segmentation algorithms at a small computational cost. 
In this paper we make use of LaserNet++ approach to fuse LiDAR and camera in RV, before combining the fused features with the BEV feature map.

\begin{figure*}[ht!]
\centering
\includegraphics[width=1.0\textwidth]{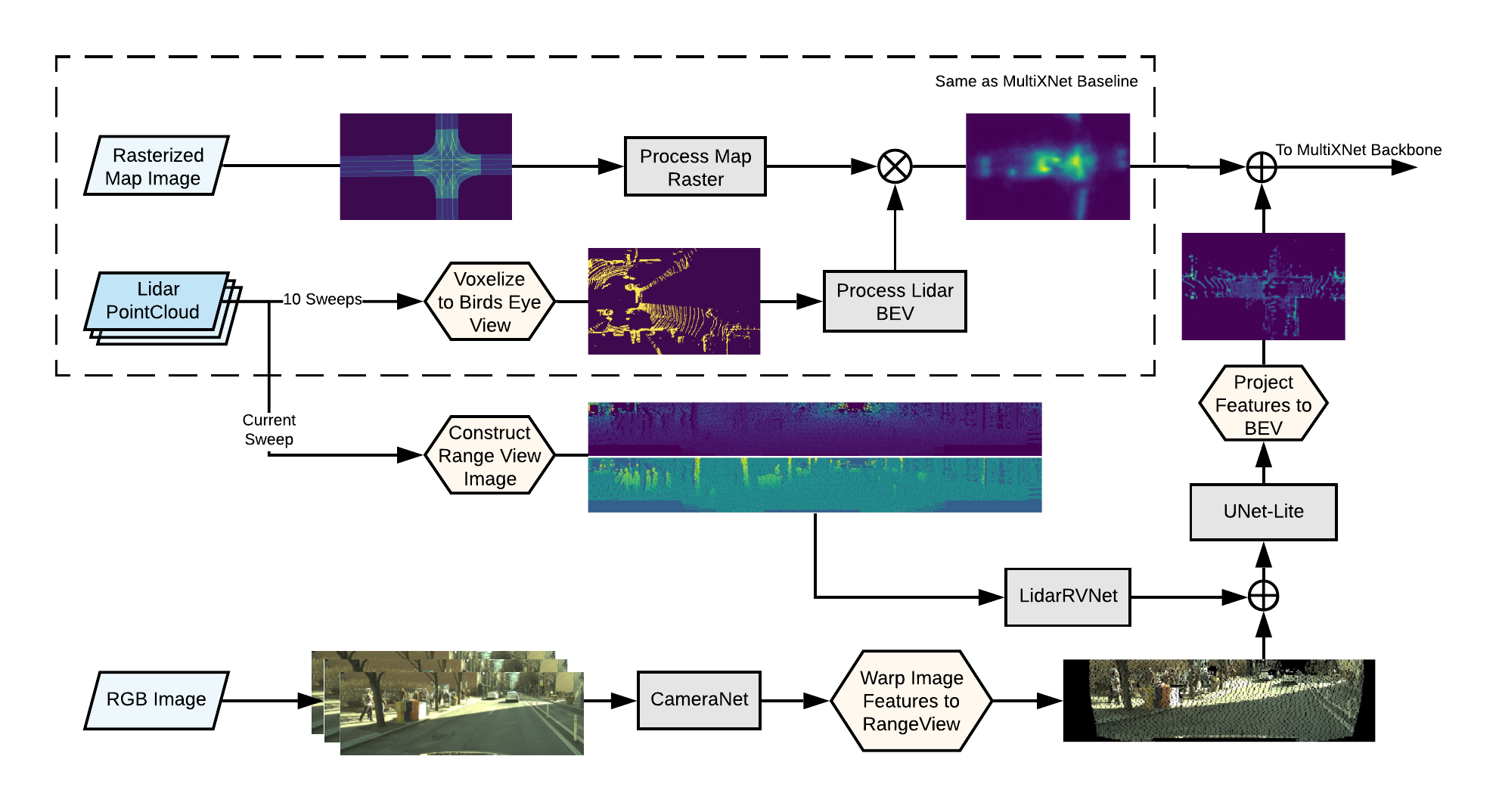}
\caption{Overview of the proposed multi-view fusion architecture, where the BEV network is enclosed in dash line; LiDAR and image inputs are further processed and fused in RV, before being projected into BEV for multi-view fusion}
\label{fig:arch}
\end{figure*}

In recent years it became increasingly common to employ a single end-to-end trained model to jointly perform object detection and trajectory prediction, an approach introduced by the authors of FaF \cite{luo2018fast}. 
A follow-up work IntentNet \cite{casas2018intentnet} further uses map information and predicts both trajectories and actors' high-level intents. SpaGNN \cite{casas2020spagnn} proposes a two-stage model where it crops the feature map for each actor with rotated ROI (RROI), and models actor interactions and uncertainties by a graph neural network (GNN).
PnPNet \cite{liang2020pnpnet} performs detection, prediction and tracking in one end-to-end network. LaserFlow \cite{meyer2020laserflow} and RVFuseNet \cite{rvfusenet} extend LaserNet with multi-sweep LiDAR inputs in RV and are able to perform both detection and prediction.
STINet \cite{zhang2020stinet} proposes a spatial-temporal network that generates detection, future trajectory, and past locations for each pedestrian, and handles actor interactions with GNN. 
MultiXNet \cite{djuric2020multixnet} extends IntentNet with second-stage trajectory refinement, joint detection and prediction on multiple classes of actors, and multi-modal trajectory prediction, and is able to achieve state-of-the-art results. 
Very recently and concurrent to this work, MVFuseNet \cite{Laddha_2021_CVPR} builds on MultiXNet with a multi-view feature extractor for LiDAR data.
To demonstrate the benefits of our proposed fusion method, we apply it to the current state-of-the-art MultiXNet model, resulting in significant boost in detection performance while keeping the additional computational cost low.
\section{PROPOSED APPROACH}
\label{sect:methodology}

In this section we present our multi-view approach to jointly detect and forecast motion for multiple actor types using raw sensor data: LiDAR point clouds, camera RGB images, and HD maps. 
We describe the model inputs and then delve into network architecture and learning objective. 

\subsection{Input representation}

\textbf{LiDAR point cloud}:
As discrete convolutions operate on grid-structured inputs, previous approaches proposed to convert the 3D point cloud into either a BEV grid or an RV grid. 
A BEV grid simplifies the problem of object detection due to preservation of metric space and size constancy. 
It also simplifies the problem of motion prediction due to effective and straightforward fusion of historical LiDAR data with HD map on the BEV grid, as shown by MultiXNet \cite{djuric2020multixnet}. 
On the other hand, an RV grid preserves detailed information about smaller objects and allows efficient fusion of other RV sensors such as camera, as shown by \cite{meyer2019lasernet} and \cite{Meyer_2019_CVPR_Workshops}. 
Unlike the earlier work, our proposed method exploits both RV and BEV representations of LiDAR data.

We collect the LiDAR returns in sweeps, each containing data collected by the sensor in one full $360\degree$ rotation. We follow the approach proposed in \cite{djuric2020multixnet} and encode a history of LiDAR sweeps in a BEV grid.
Each LiDAR sweep $\mathcal{S}_t$  at time $t$ comprises LiDAR points represented by their $(x,y,z)$ locations. 
Then, sweep $\mathcal{S}_t$ is voxelized in a BEV image centered on the SDV, with voxel sizes of $\Delta_L$, $\Delta_W$, and $\Delta_V$ along the $x$-, $y$-, and $z$-axes, respectively. 
Each voxel encodes binary occupancy indicating whether there exists at least one LiDAR point in that voxel. With this setting, the input feature for each sweep $\mathcal{S}_t$ is of shape $\left \lceil{\frac{L}{\Delta_L}}\right \rceil \times \left \lceil{\frac{W}{\Delta_W}}\right \rceil \times T\left \lceil{\frac{V}{\Delta_V}}\right \rceil$, where length $L$, width $W$, and height $V$ define the area of interest in a 3D physical space around the SDV.
We also encode the $T-1$ past sweeps $\{\mathcal{S}_{t - T + 1}, \ldots, \mathcal{S}_{t - 1}\}$ into the same BEV frame, and stack the feature maps along the channel dimension.

For RV representation of the LiDAR sweep we follow the approach from \cite{meyer2019lasernet}.
Each LiDAR point is described by range $r$, intensity $e$, azimuth $\theta$, and laser ID $m$, which maps to a known elevation angle of the corresponding laser beam.
The input RV image is generated by converting laser ID $m$ into rows and discretizing azimuth $\theta$ into columns, covering a full $360\degree$ view around the SDV. 
We pick the horizontal resolution of the RV image such that all sensor readings fall into different pixels. Each image pixel contains four channels: range $r$, height from the ground $z$, intensity $e$, and a binary value indicating whether the cell contains a valid sensor reading (for invalid sensor readings we set values of the other three channels to $-1$). In this paper we only embed the current LiDAR sweep $\mathcal{S}_t$ in RV, while past sweeps are only used in the BEV embedding described above. We feel that RV from single sweep is dense enough for detection purposes and fusing multiple sweeps in RV increases the complexity and latency in the network as shown in \cite{meyer2020laserflow}.

\textbf{High-definition map}:
Similarly to \cite{djuric2020multixnet}, we form a BEV representation of an HD map by rasterization. Specifically, we encode static map elements in the same frame as our BEV LiDAR grid. 
Static map elements include driving paths, crosswalks, lane and road boundaries, intersections, driveways, and parking lots, where each element is encoded as a binary mask in its own separate channel.
This results in a total of seven map channels used as a model input, shown in the top part of Figure \ref{fig:arch}.

\textbf{Camera image}:
We use the current RGB image frame from front facing camera as an additional input. It is expected for camera to help with object detection at longer ranges, where LiDAR data is sparse.

\subsection{Network architecture} \label{Architecture}
In this section we introduce a novel multi-view architecture, illustrated in Figure \ref{fig:arch}. 
Our architecture is composed of two major parts, feature extractors and feature projectors.
We first explain feature projector logic, and then explain how projection is used along with the feature extractor that is tasked with processing and fusing different views.

\subsubsection{Point-based feature projection}

To project features from one view to another we use a simple, generic method, described below.
In our proposed approach we use three views: camera view, LiDAR RV, and BEV. 
The projection between these views is computed as follows,
\begin{equation}
\label{eq:point_proj}
    F_{t}(x) = \frac{\sum_{i=1}^N 1_{x=\mathcal{P}_{t}(L_i)} F_{s}\big(\mathcal{P}_{s}(L_i)\big)}{\sum_{i=1}^N 1_{x=\mathcal{P}_{t}(L_i)}}.
\end{equation}
Here $F_{t}$ and $F_{s}$ are feature maps in the target and source views, respectively, represented as $2$D grids and indexed by a cell index, $x$ denotes a cell index within the target grid, scalar $N$ is the total number of LiDAR points in a sweep, while $1_{c}$ is an indicator function that equals 1 if condition $c$ holds true and 0 otherwise. 
$\mathcal{P}_{t}$ and $\mathcal{P}_{s}$ are projection operators that project LiDAR point $L_i$ onto the corresponding view and return the cell index in that view. For example, projection operator for camera $\mathcal{P}_{cam}(L_{i})$ would be $K(RL_{i} + t')$, where $K$ is a camera matrix and $R$ and $t'$ are the rotation matrix and translation vector that transform the 3D point from the LiDAR’s coordinate frame to the camera’s coordinate frame. For $\mathcal{P}_{rv}(L_{i})$, as explained in earlier section, we use laser ID  as row index and discretized azimuth $\theta$ as column. 
And similarly for $\mathcal{P}_{bev}(L_{i})$, we discretize $y$- and $x$-coordinate of lidar point $L_{i}$ to compute row and column index of the BEV grid representation, respectively. 
In other words, we use LiDAR points to extract features from the source view and project them into the target view, then apply average pooling if multiple points are projected into the same target cell.

\subsubsection{Multi-view fusion}

Our multi-view model takes in two branches of LiDAR inputs, one for BEV inputs and another for RV inputs. 
In the remainder of the text, we refer to them as BEV branch and RV branch for simplicity. 

\textbf{BEV branch:} The BEV branch is similar to MultiXNet \cite{djuric2020multixnet}, where we separately extract features from voxelized multi-sweep LiDAR data on one side and rasterized HD map on the other. 
The two sets of features are then summed, before being sent downstream to MultiXNet first-stage backbone network explained in the following section. 

\textbf{RV branch:} The RV branch consumes LiDAR point-cloud data and RGB camera images, and fuses them in the RV before projection onto BEV (note that this branch was not present in the base MultiXNet architecture, as seen in Figure \ref{fig:arch}). 
The RV camera fusion is inspired by prior work \cite{Meyer_2019_CVPR_Workshops}, which showed that fusion of camera features with the RV LiDAR features is both efficient and effective. 
This can be explained by two reasons: 
\begin{enumerate}[label=(\roman*)]
    \item Projection distortion of camera features from camera to RV LiDAR frame has less severe consequences than direct projection to BEV. While small calibration or synchronization errors may result in large BEV projection errors (especially at the object boundaries), the RV projection errors result in shifts of only a few pixels, making them recoverable even with small networks.
    \item As the projection distortion is minimal and recoverable, LiDAR and camera RV features can be fused early to share computation, improving efficiency.
\end{enumerate}

Next we discuss the RV branch in detail. 
We first construct the RV image from LiDAR point cloud, which is then processed by 2 convolutional layers with $3 \times 3$ kernel without stride. 
In parallel, we extract features from RGB camera image with a lightweight {\em CameraNet}, which consists of $6$ layers of convolutions, each with $3 \times 3$ kernel and stride $2$ for every alternate layer. Additionally, we also experiment with ResNet34 as image-module for direct apple-to-apple comparison with ContFusion \cite{liang2018deep} which uses the same network for image processing.
We project these extracted camera features to LiDAR RV using equation \eqref{eq:point_proj}, then concatenate them with the LiDAR RV features along with a binary indicator encoding whether the RV cell contains a valid camera projection (we use $1$ and $-1$ to indicate valid and invalid projection, respectively). 
Next, we apply a multi-scale U-Net \cite{ronneberger2015u} on the concatenated features. 
Our U-Net architecture processes the features at 2 scales, each down-scaling 2x horizontally and doubling the channel size. We use a residual block with skip connection as a processing block at each level, followed by a deconv layer for feature upsampling. 
Lastly, the resulting RV features are projected into BEV using equation \eqref{eq:point_proj}.

\subsubsection{Backbone network and loss function}

The rest of processing is performed in BEV frame, following a two-stage architecture proposed by MultiXNet \cite{djuric2020multixnet}. 
We do not explore other backbone architectures since MultiXNet was already shown to provide state-of-the-art performance \cite{djuric2020multixnet}.
In addition, goal of this study is to focus on the proposed multi-view fusion approach and its relative performance impact, making the investigation of various backbone architectures out of scope of our current work.

More specifically, in the first stage of the backbone network the model outputs a detection for each grid cell, outputting existence probability ${\hat p}$, bounding box center (${\hat c}_{x0}, {\hat c}_{y0}$), length ${\hat l}$ and width ${\hat w}$ of the bounding box, and heading ${\hat \theta}_0$. 
The trajectory prediction comprises waypoint centers (${\hat c}_{xh}, {\hat c}_{yh}$) and headings ${\hat \theta}_h$ at future time horizon $h$. 
Bounding box shape is considered constant across the prediction horizon.
We define foreground cells (fg) as cells containing labeled objects.
Then, loss function at prediction horizon $h$ computed at such cells is defined as follows,
\begin{align}
\begin{split}
\label{eq:loss_pred}
\mathcal{L}_{\textnormal{fg}(h)} = ~& 1_{h=0} \Big(\ell_{\textnormal{focal}}({\hat p}) + \ell_1({\hat l} - l) + \ell_1({\hat w} - w) \Big) + \\
& \ell_1({\hat c}_{xh} - c_{xh}) + \ell_1({\hat c}_{yh} - c_{yh}) + \\
& \ell_1(\sin{\hat \theta}_h - \sin{\theta}_h) + \ell_1(\cos{\hat \theta}_h - \cos{\theta}_h).
\end{split}
\end{align}
Here $\ell_{\textnormal{focal}}({\hat p}) = -(1-{\hat p})^\gamma \log{{\hat p}}$ is a focal loss \cite{lin2017focalloss} with $\gamma=2$, and $\ell_1$ is the smooth-$\ell_1$ regression loss.
Values $l$, $w$, $c_{xh}$, $c_{yh}$, and $\theta_h$ are ground-truth values. 
For cells without labeled objects called background cells (bg), the loss is $\mathcal{L}_{\textnormal{bg}} = \ell_{\textnormal{focal}}(1 - {\hat p})$. 
Then, the overall first-stage loss equals
\begin{equation}
    \mathcal{L} = 1_{\textnormal{bg cell}} \mathcal{L}_{\textnormal{bg}} + 1_{\textnormal{fg cell}} \sum_{h=0}^H \lambda^{h}  \mathcal{L}_{\textnormal{fg}(h)},
\end{equation}
where $\lambda \in (0, 1)$ is a constant decay factor used to reduce the impact of errors at longer prediction horizons (set to $0.97$ in our experiments). 

In addition, we also model location uncertainty of each waypoint, refine the trajectories for vehicles with a second-stage rotated region-of-interest cropping, and output multi-modal trajectory predictions for vehicle actors. 
We refer the reader to \cite{djuric2020multixnet} for details, which are omitted here due to limited space. 
The final output of the proposed model contains the detection bounding boxes and trajectory predictions for three main classes of road actors, namely vehicles, pedestrians, and bicyclists.
\section{EXPERIMENTS}
\label{sect:experiments}
In this section we perform empirical evaluation of the model with our proposed multi-view fusion method, and compare both quantitative and qualitative performance results with the existing baseline state-of-the-art approaches that jointly solve perception and prediction tasks.

\subsection{Data sets}
\setlength\tabcolsep{1.5pt} 
\begin{table*} 
\normalsize
\caption{Evaluation on ATG4D using detection AP (\%), prediction DE ($\text{cm}$), and latency ($\text{ms}$)}
\label{tab:ATG4D}
\centering
{
  \fontsize{7.5}{9}\selectfont
  \begin{tabularx}{\textwidth}{c
  >{\centering\arraybackslash}X 
  >{\centering\arraybackslash}X 
  >{\centering\arraybackslash}X
  >{\centering\arraybackslash}X
  >{\centering\arraybackslash}X 
  >{\centering\arraybackslash}X 
  >{\centering\arraybackslash}X 
  >{\centering\arraybackslash}X 
  >{\centering\arraybackslash}X
  >{\centering\arraybackslash}X
  >{\centering\arraybackslash}X 
  >{\centering\arraybackslash}X 
  >{\centering\arraybackslash}X 
  >{\centering\arraybackslash}X 
  >{\centering\arraybackslash}X
  >{\centering\arraybackslash}X
  >{\centering\arraybackslash}X 
  >{\centering\arraybackslash}X 
  c} 
       & \multicolumn{6}{c}{\bf Vehicles} & \multicolumn{6}{c}{\bf Pedestrians} & \multicolumn{6}{c}{\bf Bicyclists} & \\
       \cmidrule(lr){2-7} \cmidrule(lr){8-13} \cmidrule(lr){14-19}
       & \multicolumn{5}{c}{$\textbf{AP}_{0.7}$ $\uparrow$} &  & \multicolumn{5}{c}{$\textbf{AP}_{0.1}$ $\uparrow$} & & \multicolumn{5}{c}{$\textbf{AP}_{0.3}$ $\uparrow$} &  & \\
       \cmidrule(lr){2-6} \cmidrule(lr){8-12} \cmidrule(lr){14-18}
       &{\bf Full} & \multicolumn{4}{c}{\bf In Camera FOV} & \bf DE$\downarrow$& {\bf Full}&  \multicolumn{4}{c}{\bf In Camera FOV } & \bf DE$\downarrow$& {\bf Full}&  \multicolumn{4}{c}{\bf In Camera FOV}  & \bf DE$\downarrow$& \bf Lat.$\downarrow$\\
       \cmidrule{3-6} \cmidrule{9-12} \cmidrule{15-18} 
       
    \bf Method &  &  0-75m & 0-25 & 25-50 & 50-75 &  &  & 0-75m & 0-25 & 25-50 & 50-75 & &  & 0-75m & 0-25 & 25-50 & 50-75 & & \\
    
    \hline
    \rowcolor{lightgray}
    MultiXNet & \bf 84.2 & 83.9 & \bf 92.6 & 85.5 & 68.6 & 80 & 88.6 & 88.9 & 87.3 & 89.5 & 88.0 & 57 & 84.6 & 82.2 & 87.4 & 78.7 & \bf 76.8 & \bf 49 & 35.2\\
    LC-MV-Lite & \bf 84.7 & \bf  84.5 & \bf  92.6 & \bf 86.1 & \bf 70.1 & \bf 79 & \bf 89.4 & \bf 89.8 & \bf 88.5 & \bf 90.1 & \bf 88.7 & \bf 57 & \bf 87.3 & \bf 85.2 & \bf 90.8 & \bf 81.5 & \bf 76.8 & \bf 49 & 43.4\\
    \hline
\end{tabularx}
}
\end{table*}

\setlength\tabcolsep{1.5pt} 
\begin{table*}
\normalsize
\caption{Evaluation on nuScenes using detection AP (\%), prediction DE ($\text{cm}$), and latency ($\text{ms}$)}
\label{tab:nuscenes}
\centering
{
  \fontsize{7.5}{9}\selectfont
  \begin{tabularx}{\textwidth}{
  c
  >{\centering\arraybackslash}X 
  >{\centering\arraybackslash}X 
  >{\centering\arraybackslash}X 
  >{\centering\arraybackslash}X 
  >{\centering\arraybackslash}X 
  >{\centering\arraybackslash}X 
  >{\centering\arraybackslash}X 
  >{\centering\arraybackslash}X 
  >{\centering\arraybackslash}X 
  >{\centering\arraybackslash}X 
  >{\centering\arraybackslash}X 
  >{\centering\arraybackslash}X 
  >{\centering\arraybackslash}X 
  >{\centering\arraybackslash}X 
  >{\centering\arraybackslash}X 
  c}
       & \multicolumn{5}{c}{\bf Vehicles} & \multicolumn{5}{c}{\bf Pedestrians} & \multicolumn{5}{c}{\bf Bicyclists} & \\
       \cmidrule(lr){2-6} \cmidrule(lr){7-11} \cmidrule(lr){12-16}
       & \multicolumn{4}{c}{$\textbf{AP}_{0.7}$ $\uparrow$} &  & \multicolumn{4}{c}{$\textbf{AP}_{0.1}$ $\uparrow$} & & \multicolumn{4}{c}{$\textbf{AP}_{0.3}$ $\uparrow$} & & \\
       \cmidrule(lr){2-5} \cmidrule(lr){7-10} \cmidrule(lr){12-15}
       & {\bf Full} & \multicolumn{3}{c}{\bf In Camera FOV} & {\bf DE$\downarrow$}& {\bf Full} &  \multicolumn{3}{c}{\bf In Camera FOV} & {\bf DE$\downarrow$}& {\bf Full} &  \multicolumn{3}{c}{\bf In Camera FOV}  & {\bf DE$\downarrow$}&{\bf Lat.$\downarrow$}\\
       \cmidrule{3-5} \cmidrule{8-10} \cmidrule{13-15} 

    \bf Method & &  0-50m & 0-25 & 25-50 & & & 0-50m & 0-25 & 25-50 & &  & 0-50m & 0-25 & 25-50 &  &\\
    
    \hline
    \rowcolor{lightgray}
    MultiXNet & 60.6 & 60.9 & 80.0 & 43.3 & 108 & 65.5 & 63.8 & 72.9 & 54.1 & 87 & 31.7 & 32.1 & 41.4 & 25.2 & 191 & 32.3\\
    
    ContFuse & 60.9 & 61.9 & 79.4 & 46.0 & 109 & 67.2 & 71.6 & 78.0 & 64.3 & 82 & 33.5 & 44.1 & 53.5 & 36.1 & 188 & 63.6\\
    
    \rowcolor{lightgray}
    L-MV & 61.1 & 61.5 & 79.6 & 45.3 & \bf 107 & 71.0 & 70.4 & 79.1 & 60.6 & 82 & 38.2 &  38.1 & 53.3 & 25.8 & 187 & 37.4\\
    
    LC-MV-ResNet & \bf 62.4 & \bf 63.7 &  \bf 80.8 & \bf 48.6 & \bf 107 & \bf 72.6 & \bf 76.7 & \bf 83.5 & \bf 69.5 & \bf 73 & \bf 40.9 & \bf 52.3 & \bf 64.5 & \bf 43.2 & 204 & 46.2\\
    
    \rowcolor{lightgray}
    LC-MV-Lite & \bf 62.9 & \bf 63.3 & \bf  81.2 & 47.0 & \bf 107 & 71.4 & 73.1 & 80.4 & 64.6 & 80 & 39.8 & 40.4 & 53.6 & 29.5 & \bf 179 & 38.3\\
    
    \hline
\end{tabularx}
}
\end{table*}

We evaluate our method on two autonomous driving data sets, ATG4D and nuScenes \cite{caesar2020nuscenes}.
ATG4D is a proprietary data set collected using a 64-beam LiDAR with $10\text{Hz}$ sweep capture frequency, and a front camera capturing images at $1920 \times 1200$ resolution with a horizontal field of view (FOV) of $90\degree$. 
The data contains over $1$ million frames from $5{,}500$ scenes, with 3D bounding box labels at maximum range of $100$m. 
nuScenes is a public data set that uses a 32-beam LiDAR with $20\text{Hz}$ sweep capture frequency, front camera with $1600 \times 900$ resolution and a horizontal FOV of $70 \degree$. 
The data contains $1{,}000$ scenes with $390{,}000$ LiDAR sweeps.
While we also compare our proposed method to other state-of-the-art methods on the public nuScenes data, we present an additional comparison on proprietary ATG4D data as it is a much larger and richer data set thus providing statistically more significant results, especially in the case of rare classes such as pedestrians and bicyclists. 

For ATG4D, we evaluate using the same setting as MultiXNet \cite{djuric2020multixnet}. 
The BEV input uses  $L = 150\text{m}$, $W = 100\text{m}$, $V = 3.2\text{m}$, $\Delta_L = 0.16\text{m}$, $\Delta_W = 0.16\text{m}$, $\Delta_V = 0.2\text{m}$, and we use $T = 10$ sweeps to predict $H=30$ future states at $10\text{Hz}$ (thus using $1\text{s}$ of history to predict $3\text{s}$ into future).
The RV input only uses the current LiDAR sweep, with a input resolution of $2048 \times 64$.
We use the front camera RGB image synchronized with the current LiDAR sweep, after cropping 438 pixels from the top which mostly include sky. 

For nuScenes, we use same hyper-parameters and loss functions as in the ATG4D experiment, with a few changes in input representations. The BEV input uses  $L = 100\text{m}$, $W = 100\text{m}$, $V = 8\text{m}$, $\Delta_L = 0.125\text{m}$, $\Delta_W = 0.125\text{m}$, $\Delta_V = 0.2\text{m}$, and we use $T = 10$ sweeps at $20\text{Hz}$ to predict $H=30$ future states at $10\text{Hz}$ (thus using $0.5\text{s}$ of history to predict $3\text{s}$ into future).
The RV input dimension is set to $2048 \times 32$ (note that there are less rows than for ATG4D as nuScenes uses 32-beam LiDAR).
Camera images are used directly without cropping.

\subsection{Experimental setting}
Next we describe the experiments performed to evaluate our approach. 
We refer to our MV method as LC-MV when using both LiDAR and camera as inputs, and L-MV when only using lidar where the camera sub-branch from Figure \ref{fig:arch} is removed. 
We experimented with two LC-MV variants with different image processing networks: LC-MV-ResNet uses ResNet-34, while LC-MV-Lite uses a lightweight network described in Section \ref{Architecture}.
We also consider two baseline models that perform end-to-end detection and prediction using multi-sweep LiDAR: the MultiXNet model as implemented in \cite{djuric2020multixnet}, as well as Continuous Fusion (referred to as ContFuse) \cite{liang2018deep}.

Since the original ContFuse paper focuses only on detection on single-sweep LiDAR, we cannot directly compare our model to the original ContFuse implementation, and we instead reimplement ContFuse in the MultiXNet framework to ensure fair comparison. ContFuse projects the camera feature into a dense BEV feature map. 
For each target BEV pixel, it finds the nearest $K$ LiDAR points within distance $d$ and projects the LiDAR points onto the camera feature map to retrieve the corresponding image features. 
These features are concatenated with geometry information and fed into multi-layer perceptron (MLP) to output the per-pixel feature in BEV.
We extracted camera features of dimension size $32$ from ResNet-34 \cite{resnet} backbone with an extra Feature Pyramid Module \cite{DBLP:journals/corr/LinDGHHB16}. 
We set the neighbor count $K=1$ and distance $d=3\text{m}$. 
The 3D offsets between the source LiDAR point and the target BEV pixel are used as additional geometry information. 
A 3-layer MLP with hidden layer of size $128$ is used for feature extraction. 
The projected camera BEV features is added to the LiDAR+map BEV features, at the same layer as in our proposed architecture from Fig. \ref{fig:arch}.
\begin{figure*}[t]
\centering
\captionsetup{justification=centering}
\includegraphics[width=1.0\textwidth]{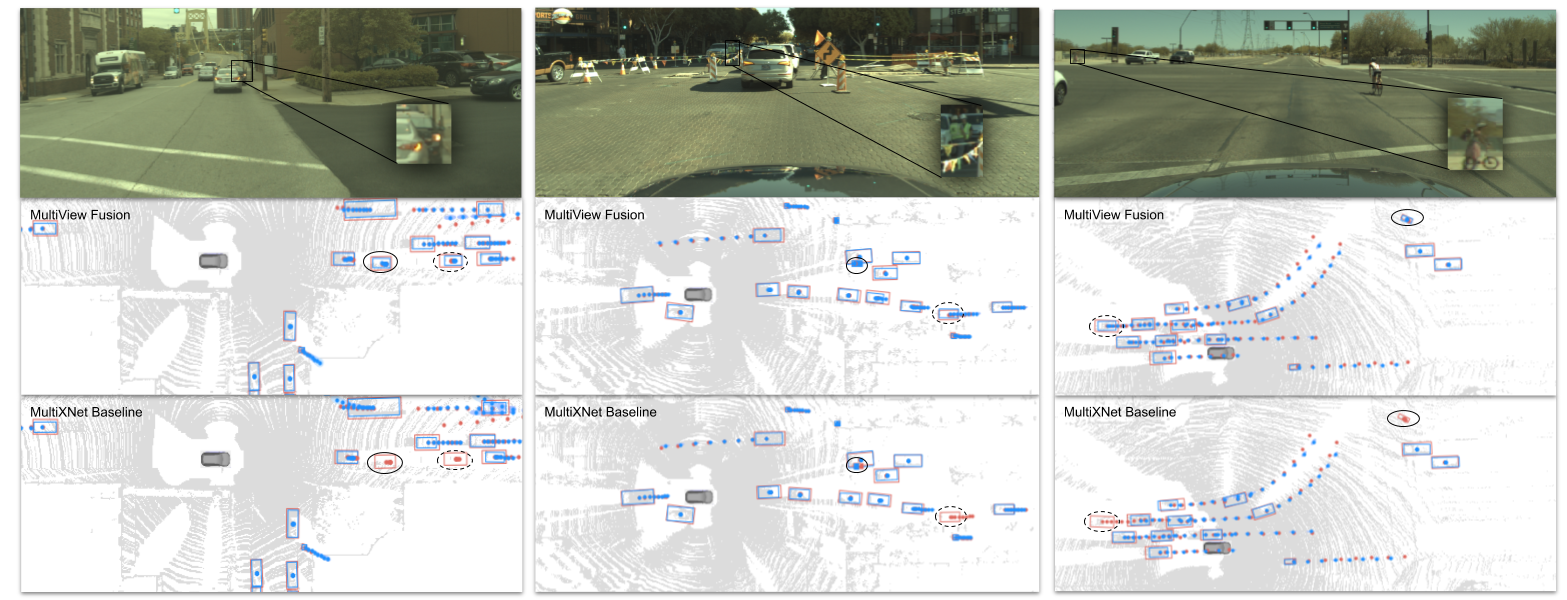}
\fontsize{7}{10}\selectfont
\caption{Qualitative comparison of LC-MV (middle) and MultiXNet (bottom) for three illustrative examples with occluded vehicle, occluded pedestrians, and occluded bicyclist, respectively; predictions are shown in blue, ground truth shown in red}
\label{fig:qualitative}
\end{figure*}

The models were trained for 2 epochs on 16 GPUs with a batch size of 2, using an Adam optimizer \cite{kingma2014adam} of an initial learning rate of 2e-4. 
The methods are evaluated following the same setting as in MultiXNet \cite{djuric2020multixnet}. 
For detection metrics, we report the Average Precision (AP) with IoU threshold set to $0.7$, $0.1$, $0.3$ for vehicles, pedestrians, and bicyclists, respectively. 
For prediction metrics, we use Displacement Error (DE) \cite{alahi2016social} evaluated at $3\text{s}$, where the detection probability threshold resulting in a recall of $0.8$ is used as the operational point.
Model inference latency is computed on RTX 2080Ti with TensorRT optimization.
Since we only use the front camera which captures a fragment of the full $360\degree$ area around SDV, we also report detection results that only include actors in camera FOV.

\subsection{Results}

We first present the model evaluation on the proprietary ATG4D data (Table \ref{tab:ATG4D}). Our proposed method (LC-MV-Lite) shows improved detection and prediction metrics across all actor classes over MultiXNet, a state-of-the-art model. The improvement is more pronounced on smaller object types (i.e., pedestrians and bicyclists) which proves that augmenting BEV methods with high resolution camera data and non-quantized RV LiDAR data is really important for detecting such small objects. The inference latency increases slightly from $35\text{ms}$ to $43\text{ms}$, which is well within the real-time requirement of typical SDV applications. 

We perform additional ablation comparisons on the public nuScenes data (Table \ref{tab:nuscenes}). 
Note that the bicyclists are one of the rarest types in nuScenes, therefore its detection AP is significantly lower than that of the other classes \cite{caesar2020nuscenes}.
When compared to the MultiXNet baseline, our LiDAR-only MV method already gives significantly improved detection AP for all three actor classes, resulting in $0.5\%$ improvement for vehicles, $5.5\%$ for pedestrians, and $6.5\%$ for bicyclists, as well as better prediction errors. 
Including camera input in the MV fusion gives further metrics improvements.
In comparison, we can see that the ContFuse baseline, while showing notable improvements over the MultiXNet baseline, performs worse when compared to the proposed LC-MV fusion models in overall metrics and has significantly higher latency.
Comparing the two LC-MV variants, LC-MV-ResNet in general performs better than LC-MV-Lite, except for having higher bicyclists prediction DE, at an expense of higher latency due to the heavy computation in ResNet-34.
Our method also performs significantly better than LaserFlow \cite{meyer2020laserflow}, a recent pure-RV joint detection and prediction model, considering the reported results on the vehicle class on nuScenes (AP: 56.1\% vs. 62.9\%, DE@3s: 143cm vs. 107cm, when comparing LaserFlow vs. LC-MV-Lite, respectively; metrics are evaluated with the same definition). 
LaserFlow does not report metrics on other actor classes on nuScenes.
Lastly, while LC-MV-ResNet achieved best prediction results for vehicle and pedestrian classes, it had degraded DE on rare bicyclist class despite the best detection results. 
This can be explained by a small bicyclist count in the data set, which resulted in somewhat unstable performance for this actor type.

To further understand the effect of camera fusion, Table \ref{tab:nuscenes} also includes detection metrics within the camera FOV, with further metrics slicing based on the actor distance to SDV in meters.
As expected, the metrics gains between L-MV and LC-MV are more pronounced within the camera FOV, as opposed to the overall metrics across the full $360\degree$ view.
With further distance-based metrics slicing, we found that the MV fusion in general shows similar or larger improvements at longer ranges. 
This is most evident for vehicle actors, yielding $0.8\%$ improvement in near range and $5.3\%$ improvement at long range when comparing MultiXNet to LC-MV-ResNet model.
The result is aligned with findings by previously reported results on RV-based detectors \cite{Meyer_2019_CVPR_Workshops,meyer2019lasernet}, indicating that the RV representation is more effective in capturing longer-range objects.


The model inference latency on nuScenes is presented in Table \ref{tab:nuscenes}. 
We can see that adding MV LiDAR fusion results in additional $5$ms inference time over the MultiXNet baseline. 
Including the front camera in our architecture further costs an additional $9$ms with LC-MV-ResNet and only $1$ms with LC-MV-Lite. 
In comparison, adding camera input to MultiXNet through ContFuse increases the latency by $31$ms, much larger slowdown than by using the proposed LC-MV models. 
Our proposed LC-MV-Lite model gives improved detection and prediction metric over MultiXNet and ContFuse, while being highly computationally efficient.
As a result, the method can easily be scaled to use multiple cameras (e.g., side and rear cameras) while maintaining real-time inference speeds.

Lastly, we present several qualitative examples illustrating how the introduced fusion method improves object detection of traffic actors in the ATG4D data set.
Figure \ref{fig:qualitative} shows three examples where the MultiXNet baseline fails to detect actors at longer distances while having a clear view in the camera image (shown as solid circles in BEV). 
The LC-MV model also detects several additional actors missed by MultiXNet, which are shown as dotted circles.
We can see that the trajectory predictions for the additional detected actors are also accurate compared to the ground truth, exemplifying good performance of the proposed approach in terms of both object detection and motion prediction.
\section{CONCLUSION}
\label{sect:conclusion}

We focused on object detection and motion prediction of traffic actors, which are safety-critical problems in the development and deployment of SDVs. 
The end-to-end approaches can be roughly grouped into BEV-based and RV-based, depending on how the input data is encoded and processed.
We recognized the advantages and disadvantages of both methods, and proposed a novel multi-view  fusion method that combines the best of both groups in a unified architecture.
We evaluated the approach on large-scale, real-world autonomous driving data, and compared to the state-of-the-art end-to-end approaches with the results strongly suggesting practical benefits of the proposed multi-view approach.
Moreover, while we applied the proposed multi-view method to MultiXNet, it is very general and is directly applicable to other BEV or RV algorithms, where it can be used to improve detection and prediction performance.

\balance
{\small
\bibliographystyle{ieee_fullname}
\bibliography{egbib}

\begin{thebibliography}{10}\itemsep=-1pt

\bibitem{alahi2016social}
Alexandre Alahi, Kratarth Goel, Vignesh Ramanathan, Alexandre Robicquet, Li
  Fei-Fei, and Silvio Savarese.
\newblock Social lstm: Human trajectory prediction in crowded spaces.
\newblock In {\em Proceedings of the IEEE conference on computer vision and
  pattern recognition}, pages 961--971, 2016.

\bibitem{caesar2020nuscenes}
Holger Caesar, Varun Bankiti, Alex~H Lang, Sourabh Vora, Venice~Erin Liong,
  Qiang Xu, Anush Krishnan, Yu Pan, Giancarlo Baldan, and Oscar Beijbom.
\newblock nuscenes: A multimodal dataset for autonomous driving.
\newblock In {\em Proceedings of the IEEE/CVF Conference on Computer Vision and
  Pattern Recognition}, pages 11621--11631, 2020.

\bibitem{casas2020spagnn}
Sergio Casas, Cole Gulino, Renjie Liao, and Raquel Urtasun.
\newblock Spagnn: Spatially-aware graph neural networks for relational behavior
  forecasting from sensor data.
\newblock In {\em 2020 IEEE International Conference on Robotics and Automation
  (ICRA)}, pages 9491--9497. IEEE, 2020.

\bibitem{casas2018intentnet}
Sergio Casas, Wenjie Luo, and Raquel Urtasun.
\newblock Intentnet: Learning to predict intention from raw sensor data.
\newblock In {\em Conference on Robot Learning}, pages 947--956, 2018.

\bibitem{chen2017multi}
Xiaozhi Chen, Huimin Ma, Ji Wan, Bo Li, and Tian Xia.
\newblock Multi-view 3d object detection network for autonomous driving.
\newblock In {\em Proceedings of the IEEE Conference on Computer Vision and
  Pattern Recognition}, pages 1907--1915, 2017.

\bibitem{cui2019multimodal}
Henggang Cui, Vladan Radosavljevic, Fang-Chieh Chou, Tsung-Han Lin, Thi Nguyen,
  Tzu-Kuo Huang, Jeff Schneider, and Nemanja Djuric.
\newblock Multimodal trajectory predictions for autonomous driving using deep
  convolutional networks.
\newblock In {\em 2019 International Conference on Robotics and Automation
  (ICRA)}, pages 2090--2096. IEEE, 2019.

\bibitem{djuric2020multixnet}
Nemanja Djuric, Henggang Cui, Zhaoen Su, Shangxuan Wu, Huahua Wang, Fang-Chieh
  Chou, Luisa~San Martin, Song Feng, Rui Hu, Yang Xu, et~al.
\newblock Multixnet: Multiclass multistage multimodal motion prediction.
\newblock In {\em IEEE Intelligent Vehicles Symposium (IV)}, 2021.

\bibitem{djuric2020}
Nemanja Djuric, Vladan Radosavljevic, Henggang Cui, Thi Nguyen, Fang-Chieh
  Chou, Tsung-Han Lin, Nitin Singh, and Jeff Schneider.
\newblock Uncertainty-aware short-term motion prediction of traffic actors for
  autonomous driving.
\newblock In {\em Proceedings of the IEEE/CVF Winter Conference on Applications
  of Computer Vision}, pages 2095--2104, 2020.

\bibitem{LueFan2021rv}
Lue Fan, Xuan Xiong, Feng Wang, Naiyan Wang, and Zhaoxiang Zhang.
\newblock Rangedet: In defense of range view for lidar-based 3d object
  detection.
\newblock {\em arXiv preprint arXiv:2103.10039}, 2021.

\bibitem{feng2020deep}
Di Feng, Christian Haase-Sch{\"u}tz, Lars Rosenbaum, Heinz Hertlein, Claudius
  Glaeser, Fabian Timm, Werner Wiesbeck, and Klaus Dietmayer.
\newblock Deep multi-modal object detection and semantic segmentation for
  autonomous driving: Datasets, methods, and challenges.
\newblock {\em IEEE Transactions on Intelligent Transportation Systems}, 2020.

\bibitem{guo2020deep}
Yulan Guo, Hanyun Wang, Qingyong Hu, Hao Liu, Li Liu, and Mohammed Bennamoun.
\newblock Deep learning for 3d point clouds: A survey.
\newblock {\em IEEE Transactions on Pattern Analysis and Machine Intelligence},
  2020.

\bibitem{resnet}
Kaiming He, Xiangyu Zhang, Shaoqing Ren, and Jian Sun.
\newblock Deep residual learning for image recognition.
\newblock In {\em Proceedings of the IEEE conference on computer vision and
  pattern recognition}, pages 770--778, 2016.

\bibitem{kingma2014adam}
Diederik~P Kingma and Jimmy Ba.
\newblock Adam: A method for stochastic optimization.
\newblock {\em arXiv preprint arXiv:1412.6980}, 2014.

\bibitem{ku2018joint}
Jason Ku, Melissa Mozifian, Jungwook Lee, Ali Harakeh, and Steven~L Waslander.
\newblock Joint 3d proposal generation and object detection from view
  aggregation.
\newblock In {\em 2018 IEEE/RSJ International Conference on Intelligent Robots
  and Systems (IROS)}, pages 1--8. IEEE, 2018.

\bibitem{rvfusenet}
Ankit Laddha, Shivam Gautam, Gregory~P Meyer, and Carlos Vallespi-Gonzalez.
\newblock Rv-fusenet: Range view based fusion of time-series lidar data for
  joint 3d object detection and motion forecasting.
\newblock {\em arXiv preprint arXiv:2005.10863}, 2020.

\bibitem{Laddha_2021_CVPR}
Ankit Laddha, Shivam Gautam, Stefan Palombo, Shreyash Pandey, and Carlos
  Vallespi-Gonzalez.
\newblock Mvfusenet: Improving end-to-end object detection and motion
  forecasting through multi-view fusion of lidar data.
\newblock In {\em Proceedings of the IEEE/CVF Conference on Computer Vision and
  Pattern Recognition (CVPR) Workshops}, pages 2865--2874, June 2021.

\bibitem{lang2019pointpillars}
Alex~H Lang, Sourabh Vora, Holger Caesar, Lubing Zhou, Jiong Yang, and Oscar
  Beijbom.
\newblock Pointpillars: Fast encoders for object detection from point clouds.
\newblock In {\em Proceedings of the IEEE Conference on Computer Vision and
  Pattern Recognition}, pages 12697--12705, 2019.

\bibitem{liang2019multi}
Ming Liang, Bin Yang, Yun Chen, Rui Hu, and Raquel Urtasun.
\newblock Multi-task multi-sensor fusion for 3d object detection.
\newblock In {\em Proceedings of the IEEE Conference on Computer Vision and
  Pattern Recognition}, pages 7345--7353, 2019.

\bibitem{liang2018deep}
Ming Liang, Bin Yang, Shenlong Wang, and Raquel Urtasun.
\newblock Deep continuous fusion for multi-sensor 3d object detection.
\newblock In {\em Proceedings of the European Conference on Computer Vision
  (ECCV)}, pages 641--656, 2018.

\bibitem{liang2020pnpnet}
Ming Liang, Bin Yang, Wenyuan Zeng, Yun Chen, Rui Hu, Sergio Casas, and Raquel
  Urtasun.
\newblock Pnpnet: End-to-end perception and prediction with tracking in the
  loop.
\newblock In {\em Proceedings of the IEEE/CVF Conference on Computer Vision and
  Pattern Recognition}, pages 11553--11562, 2020.

\bibitem{DBLP:journals/corr/LinDGHHB16}
Tsung-Yi Lin, Piotr Doll{\'a}r, Ross Girshick, Kaiming He, Bharath Hariharan,
  and Serge Belongie.
\newblock Feature pyramid networks for object detection.
\newblock In {\em Proceedings of the IEEE conference on computer vision and
  pattern recognition}, pages 2117--2125, 2017.

\bibitem{lin2017focalloss}
Tsung-Yi Lin, Priya Goyal, Ross Girshick, Kaiming He, and Piotr Doll{\'a}r.
\newblock Focal loss for dense object detection.
\newblock In {\em Proceedings of the IEEE international conference on computer
  vision}, pages 2980--2988, 2017.

\bibitem{luo2018fast}
Wenjie Luo, Bin Yang, and Raquel Urtasun.
\newblock Fast and furious: Real time end-to-end 3d detection, tracking and
  motion forecasting with a single convolutional net.
\newblock In {\em Proceedings of the IEEE conference on Computer Vision and
  Pattern Recognition}, pages 3569--3577, 2018.

\bibitem{Meyer_2019_CVPR_Workshops}
Gregory~P. Meyer, Jake Charland, Darshan Hegde, Ankit Laddha, and Carlos
  Vallespi-Gonzalez.
\newblock Sensor fusion for joint 3d object detection and semantic
  segmentation.
\newblock In {\em The IEEE Conference on Computer Vision and Pattern
  Recognition (CVPR) Workshops}, June 2019.

\bibitem{meyer2020laserflow}
Gregory~P Meyer, Jake Charland, Shreyash Pandey, Ankit Laddha, Shivam Gautam,
  Carlos Vallespi-Gonzalez, and Carl~K Wellington.
\newblock Laserflow: Efficient and probabilistic object detection and motion
  forecasting.
\newblock {\em IEEE Robotics and Automation Letters}, 6(2):526--533, 2020.

\bibitem{meyer2019lasernet}
Gregory~P Meyer, Ankit Laddha, Eric Kee, Carlos Vallespi-Gonzalez, and Carl~K
  Wellington.
\newblock Lasernet: An efficient probabilistic 3d object detector for
  autonomous driving.
\newblock In {\em Proceedings of the IEEE Conference on Computer Vision and
  Pattern Recognition}, pages 12677--12686, 2019.

\bibitem{ngiam2019starnet}
Jiquan Ngiam, Benjamin Caine, Wei Han, Brandon Yang, Yuning Chai, Pei Sun, Yin
  Zhou, Xi Yi, Ouais Alsharif, Patrick Nguyen, et~al.
\newblock Starnet: Targeted computation for object detection in point clouds.
\newblock {\em arXiv preprint arXiv:1908.11069}, 2019.

\bibitem{qi2018frustum}
Charles~R Qi, Wei Liu, Chenxia Wu, Hao Su, and Leonidas~J Guibas.
\newblock Frustum pointnets for 3d object detection from rgb-d data.
\newblock In {\em Proceedings of the IEEE conference on computer vision and
  pattern recognition}, pages 918--927, 2018.

\bibitem{Charles_2017}
Charles~R Qi, Hao Su, Kaichun Mo, and Leonidas~J Guibas.
\newblock Pointnet: Deep learning on point sets for 3d classification and
  segmentation.
\newblock In {\em Proceedings of the IEEE conference on computer vision and
  pattern recognition}, pages 652--660, 2017.

\bibitem{ronneberger2015u}
Olaf Ronneberger, Philipp Fischer, and Thomas Brox.
\newblock U-net: Convolutional networks for biomedical image segmentation.
\newblock In {\em International Conference on Medical image computing and
  computer-assisted intervention}, pages 234--241. Springer, 2015.

\bibitem{pmlr-v87-yang18b}
Bin Yang, Ming Liang, and Raquel Urtasun.
\newblock Hdnet: Exploiting hd maps for 3d object detection.
\newblock In {\em Conference on Robot Learning}, volume~87, pages 146--155,
  2018.

\bibitem{yang2018pixor}
Bin Yang, Wenjie Luo, and Raquel Urtasun.
\newblock Pixor: Real-time 3d object detection from point clouds.
\newblock In {\em Proceedings of the IEEE conference on Computer Vision and
  Pattern Recognition}, pages 7652--7660, 2018.

\bibitem{zhang2020stinet}
Zhishuai Zhang, Jiyang Gao, Junhua Mao, Yukai Liu, Dragomir Anguelov, and
  Congcong Li.
\newblock Stinet: Spatio-temporal-interactive network for pedestrian detection
  and trajectory prediction.
\newblock In {\em Proceedings of the IEEE/CVF Conference on Computer Vision and
  Pattern Recognition}, pages 11346--11355, 2020.

\bibitem{zhou2019end}
Yin Zhou, Pei Sun, Yu Zhang, Dragomir Anguelov, Jiyang Gao, Tom Ouyang, James
  Guo, Jiquan Ngiam, and Vijay Vasudevan.
\newblock End-to-end multi-view fusion for 3d object detection in lidar point
  clouds.
\newblock In {\em Conference on Robot Learning}, pages 923--932. PMLR, 2020.

\bibitem{zhou2018voxelnet}
Yin Zhou and Oncel Tuzel.
\newblock Voxelnet: End-to-end learning for point cloud based 3d object
  detection.
\newblock In {\em Proceedings of the IEEE Conference on Computer Vision and
  Pattern Recognition}, pages 4490--4499, 2018.

\bibitem{Bertha2015}
Julius Ziegler, Philipp Bender, Markus Schreiber, et~al.
\newblock Making bertha drive—an autonomous journey on a historic route.
\newblock {\em IEEE Intelligent Transportation Systems Magazine}, 6:8--20, 10
  2015.

\end{thebibliography}
}

\end{document}